\pdfoutput=1
 \documentclass[pmlr,twocolumn,10pt]{jmlr} 

\usepackage{graphicx}
\usepackage[export]{adjustbox}
\usepackage{float}
\usepackage{amsmath}
\usepackage{bm}
\usepackage{multirow}
\usepackage[normalem]{ulem}
\useunder{\uline}{\ul}{}

\usepackage{booktabs}
\usepackage{siunitx}

\usepackage[switch]{lineno}

\theorembodyfont{\upshape}
\theoremheaderfont{\scshape}
\theorempostheader{:}
\theoremsep{\newline}

\jmlrvolume{LEAVE UNSET}
\jmlryear{2024}
\jmlrsubmitted{LEAVE UNSET}
\jmlrpublished{LEAVE UNSET}
\jmlrworkshop{Machine Learning for Health (ML4H) 2024} 

 \title[MaskMedPaint: Masked Medical Image Inpainting]{MaskMedPaint: Masked Medical Image Inpainting with Diffusion Models for Mitigation of Spurious Correlations}

 \author{
  \Name{Qixuan Jin}\Email{qixuanj@mit.edu}\\
  \Name{Walter Gerych}\Email{wgerych@mit.edu}\\
  \Name{Marzyeh Ghassemi}\Email{mghassem@mit.edu}\\
  \addr Massachusetts Institute of Technology, Cambridge, US
 }

\begin{document}

\maketitle

\begin{abstract}
Spurious features associated with class labels can lead image classifiers to rely on shortcuts that don't generalize well to new domains. This is especially problematic in medical settings, where biased models fail when applied to different hospitals or systems. In such cases, data-driven methods to reduce spurious correlations are preferred, as clinicians can directly validate the modified images. While Denoising Diffusion Probabilistic Models (Diffusion Models) show promise for natural images, they are impractical for medical use due to the difficulty of describing spurious medical features. To address this, we propose Masked Medical Image Inpainting (MaskMedPaint), which uses text-to-image diffusion models to augment training images by inpainting areas outside key classification regions to match the target domain. We demonstrate that MaskMedPaint enhances generalization to target domains across both natural (Waterbirds, iWildCam) and medical (ISIC 2018, Chest X-ray) datasets, given limited unlabeled target images.
\end{abstract}

\paragraph*{Data and Code Availability}
All datasets are publicly accessible. Researchers can find instructions for downloading the \href{https://github.com/kohpangwei/group_DRO}{Waterbirds} dataset, the \href{https://github.com/p-lambda/wilds}{iWildCam} dataset, the \href{https://challenge.isic-archive.com/data/#2018}{ISIC 2018} dataset, and the two chest X-ray datasets (\href{https://physionet.org/content/mimic-cxr-jpg/2.0.0/}{MIMIC-CXR} and \href{https://www.kaggle.com/datasets/nih-chest-xrays/data}{NIH Chest X-ray}) through the corresponding hyperlinks. The specific splits of our spurious ISIC and iWildCam datasets will be released upon publication. Our GitHub repo is available at: \url{https://github.com/QixuanJin99/generative_validation}.  

\paragraph*{Institutional Review Board (IRB)}
Our research does not require IRB approval. 

\section{Introduction}
\label{sec:intro} 
Spurious features are features that correlate with the target class but are not essential for its prediction \citep{rosenfeld2018elephant, gulrajani2020search}. 
Generalizability is essential in medical settings, to allow for manual model evaluation by healthcare professionals before deployment \citep{vellido2020importance, salahuddin2022transparency}. 
Data augmentation methods are preferred over model-based methods in medical settings, because the edited images can be easily reviewed and validated by clinicians \citep{rasheed2022explainable, kelly2019key}. 


\begin{figure*}[thbp]
    \centering
    \includegraphics[width=0.8\linewidth,valign=m]{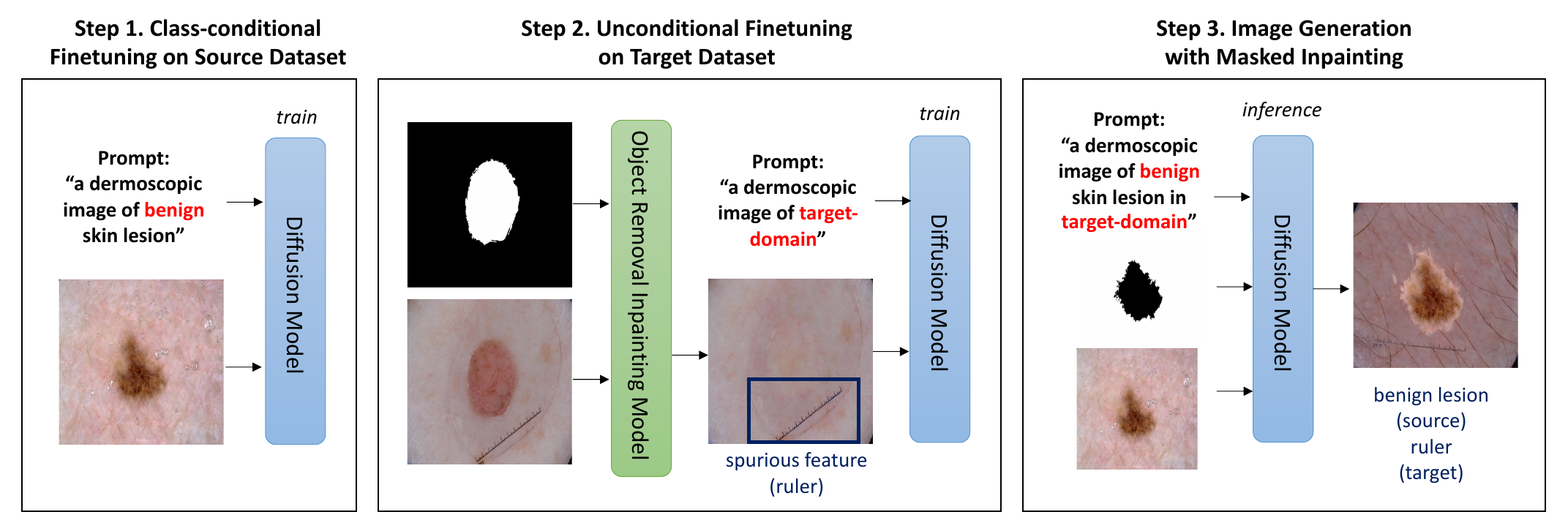}
    \caption{\textbf{MaskMedPaint pipeline.} 
    In step 1, the text-to-image diffusion model is finetuned on the source dataset $D_{source}$, with the class names formatted directly in the text prompts. The diffusion model learns class-conditional features (e.g. benign versus malignant). In step 2, we preprocess the target images by segmenting the ROI and removing the ROI with an inpainting model. We obtain background images of the target domain, which we then pair with a dummy token (``target-domain'') in the text prompt and finetune the model from Step 1 with Dreambooth. In step 3, we transfer the source images to the style of the target domain by protecting the ROI with the segmentation mask and using the model from step 2 to inpaint the remaining regions. We condition on both the class name and the dummy token in the text prompt.}
    \label{fig:main_pipeline}
\end{figure*}
 
In this work, we focus on a realistic medical setting where source data is larger and labeled, while target data (e.g. another hospital, rare disease subpopulation) is unlabeled, limited in size, and may be shifted~\citep{willemink2020preparing, bonevski2014reaching}. We propose Masked Medical Image Inpainting (MaskMedPaint), a diffusion-based method that uses masked inpainting to align source images with the target domain style (Figure \ref{fig:main_pipeline}). 


\section{Related Work}
\subsection{Shortcut Learning}
Methods that seek to mitigate the effect of shortcut learning fall under a few general approaches: sample reweighting \citep{nam2020learning, zhou2022model}, data augmentation \citep{han2022umix, yun2019cutmix, yao2022lisa}, retraining model components \citep{kirichenko2022last, moayeri2024spuriosity}, adversarial robustness \citep{ganin2015unsupervised, ganin2016domain, xie2017controllable, li2018deep}. 
In this work, we focus on data augmentation methods. We view the other methods as complementary to data augmentation, as data augmentation can always be combined with another model-based method.  

Data augmentation methods such as CutMix \citep{yun2019cutmix}, Mixup \citep{zhang2017mixup}, and LISA \citep{yao2022lisa} have applied sample interpolation and non-generative image editing techniques to improve the generalization of pretrained classifiers. More recently, text-to-image generative approaches have been used for more complex data augmentation through direct language-guided image editing. 

In \citet{dunlap2022using} specifically, Latent Augmentation using Domain Descriptions  (LADS) utilizes the dual visual-text embedding space of a pretrained CLIP model to perform transfer based on an user-defined keyword of the target domain. In \citet{dunlap2024diversify}, Automatic Language-guided Image Augmentation (ALIA) utilizes pretrained image-captioning models to automately construct text prompts used for target domain augmentation. 
 
\subsection{Spurious Features in Medical Imaging Domain}
Spurious features have been found in various medical imaging applications. In chest X-rays, spurious features like chest drains \citep{oakden2020hidden} and metal artifacts \citep{zech2018variable} negatively impact model performance across disease subgroups and datasets \citep{degrave2021ai, pooch2020can}. In dermatology images, annotations such as marker markings and reference rulers are more frequently seen with skin lesions with melanoma than with healthy lesions \citep{winkler2019association, nauta2021uncovering}. 

\subsection{Denoising Diffusion Probabilistic Models}
Denoising diffusion probabilistic models (diffusion models) demonstrate state-of-the-art performance for text-to-image generation \citep{dhariwal2021diffusion}. In particular, Stable Diffusion \citep{rombach2022high} has been widely used for art creation \citep{liao2022artbench} and image editing \citep{yang2023paint, brooks2023instructpix2pix}. Methods such as Textual Inversion \citep{gal2022image} and Dreambooth \citep{ruiz2023dreambooth} allow pretrained diffusion models to learn specific, new concepts given a few sample images. 

\section{Masked Medical Inpainting (MaskMedPaint)}
\paragraph{Class-conditional Source Finetuning}
To encapsulate image features of the prediction classes, we first finetune the diffusion model on the labeled source images. Class labels are converted into text prompts (Appendix \ref{appendix:prompt}).

\paragraph{Unconditional Target Finetuning}
Next, we use Dreambooth \citep{ruiz2023dreambooth} to finetune the diffusion model on the unlabeled target images, cleaned to be just the background with LaMa \citep{suvorov2022resolution}. Dreambooth personalizes pretrained text-to-image models to associate specific visual concepts with particular text tokens given a limited set of example images. We define a dummy token (e.g. ``target-domain'') to represent the general style of the target domain backgrounds. 

\paragraph{Masked Inpainting Image Generation}
We augment the source training images by masking the ROI, and use the Dreambooth-trained diffusion model to inpaint the rest of the image. We use text prompts with both the class name and the dummy target token, such as ``a photo of landbird with target background.'' The classifier is then trained on the combined set of original and augmented source images. 

\section{Experimental Setup}
Results for natural image datasets Waterbirds \citep{sagawa2019distributionally} and iWildCam 2020 \citep{koh2021wilds} are in the appendix. 

\subsection{Datasets}
International Skin Imaging Collaboration (ISIC) 2018 \citep{codella2019skin} is a dataset of dermoscopic skin images that are labeled for skin conditions such as melanoma. We construct a subset version with rulers as the spurious feature. MIMIC-CXR \citep{johnson2019mimic} and NIH ChestXray14 \citep{wang2017hospital} are the two chest X-ray datasets we use for the transfer. 

\subsection{Baselines}
We compare our method against a classifier trained without any data augmentation (\textbf{Base}), and a classifier trained on images with only the ROI, and the remaining area masked out (\textbf{Masked}). We also include traditional data augmentation baselines \textbf{CutMix} \citep{yun2019cutmix} and \textbf{Mixup} \citep{zhang2017mixup} with built-in torchvision transforms. We compare against \textbf{ALIA} \citep{dunlap2024diversify} for the closest diffusion-based baseline. In ALIA, we use the default image captioning and LLM summarization models provided. We compare with \textbf{LADS} \citep{dunlap2022using} for a different, CLIP-based methodology.  

\subsection{Implementation}
We use the HuggingFace \citep{von-platen-etal-2022-diffusers} implementation of Stable Diffusion 1.5 \citep{rombach2022high} for all diffusion-related methods (``runwayml/stable-diffusion-v1-5''). We vary the strengths $\in \{0.5, 0.7, 0.9, 1.0\}$ and guidance scale $\in \{7.5, 15, 20 \}$. 
For the image classifier, we use DenseNet121 model \citep{huang2017densely} from the Torchvision library \citep{torchvision2016}. All classifiers are initialized with ImageNet weights. We train with the Adam optimizer \citep{kingma2014adam} at a learning rate of $1\mathrm{e}{-3}$, a weight decay of $1\mathrm{e}{-4}$, and a batch size of 64. All images are resized to dimensions 224$\times$224 and normalized to the ImageNet distribution. 

\section{Experiments}
\subsection{Mitigating Artifact Bias in ISIC Dermoscopic Images}
\begin{table*}[t]
\centering
\resizebox{0.7\textwidth}{!}{%
\begin{tabular}{ll|rrr|rrr}
\toprule
\textbf{Dataset} & \textbf{Method} & \multicolumn{3}{c}{\textbf{Source Test Accuracy}} & \multicolumn{3}{c}{\textbf{Target Test Accuracy}} \\
\midrule
 & \multicolumn{1}{c}{\textbf{}} & \multicolumn{1}{c}{\textbf{Mean}} & \multicolumn{1}{c}{\textbf{\begin{tabular}[c]{@{}c@{}}Lower \\ 95 CI\end{tabular}}} & \multicolumn{1}{c}{\textbf{\begin{tabular}[c]{@{}c@{}}Upper \\ 95 CI\end{tabular}}} & \multicolumn{1}{c}{\textbf{Mean}} & \multicolumn{1}{c}{\textbf{\begin{tabular}[c]{@{}c@{}}Lower \\ 95 CI\end{tabular}}} & \multicolumn{1}{c}{\textbf{\begin{tabular}[c]{@{}c@{}}Upper \\ 95 CI\end{tabular}}} \\
 \midrule
\multirow{7}{*}{\textbf{ISIC 2018}} & Base & 0.930 & 0.889 & 0.972 & 0.146 & 0.073 & 0.218 \\
 & CutMix & 0.925 & 0.914 & 0.937 & 0.147 & 0.119 & 0.175 \\
 & Mixup & {\ul \textbf{0.948}} & {\ul \textbf{0.939}} & {\ul \textbf{0.957}} & 0.123 & 0.104 & 0.142 \\
 & Masked & 0.671 & 0.527 & 0.815 & {\ul \textbf{0.385}} & {\ul \textbf{0.286}} & {\ul \textbf{0.485}} \\
 & ALIA* & \textbf{0.940} & \textbf{0.922} & \textbf{0.958} & 0.106 & 0.083 & 0.129 \\
 & LADS* & 0.883 & 0.864 & 0.901 & 0.183 & 0.171 & 0.195 \\
 & MaskMedPaint & 0.746 & 0.690 & 0.802 & \textbf{0.344} & \textbf{0.302} & \textbf{0.386} \\
 \midrule
\textbf{Dataset} & \textbf{Method} & \multicolumn{3}{l}{\textbf{Source Test AUROC}} & \multicolumn{3}{l}{\textbf{Target Test AUROC}} \\
\midrule
\multirow{7}{*}{\textbf{CXR}} & Base & \textbf{0.714} & \textbf{0.711} & \textbf{0.717} & 0.490 & 0.480 & 0.499 \\
 & CutMix & {\ul \textbf{0.716}} & {\ul \textbf{0.712}} & {\ul \textbf{0.721}} & 0.448 & 0.431 & 0.465 \\
 & Mixup & 0.702 & 0.699 & 0.705 & \textbf{0.524} & \textbf{0.514} & \textbf{0.535} \\
 & Masked & 0.651 & 0.624 & 0.678 & 0.459 & 0.437 & 0.480 \\
 & ALIA* & 0.701 & 0.699 & 0.704 & 0.498 & 0.489 & 0.507 \\
 & LADS* & 0.703 & 0.692 & 0.714 & 0.521 & 0.483 & 0.559 \\
 & MaskMedPaint & 0.706 & 0.704 & 0.707 & {\ul \textbf{0.546}} & {\ul \textbf{0.537}} & {\ul \textbf{0.554}} \\
 \bottomrule
\end{tabular}
}
\caption{Performance metrics evaluated across the source and target domains for the medical imaging datasets. Best-performing method is bolded and underlined, and the second-best is bolded if its CI overlaps with the CI of the best-performing. Methods requiring additional pretrained models are highlighted with *.}
\label{table:medical_results}
\end{table*}

\begin{figure}[ht!]
    \centering
    \includegraphics[width=\linewidth, valign=m]{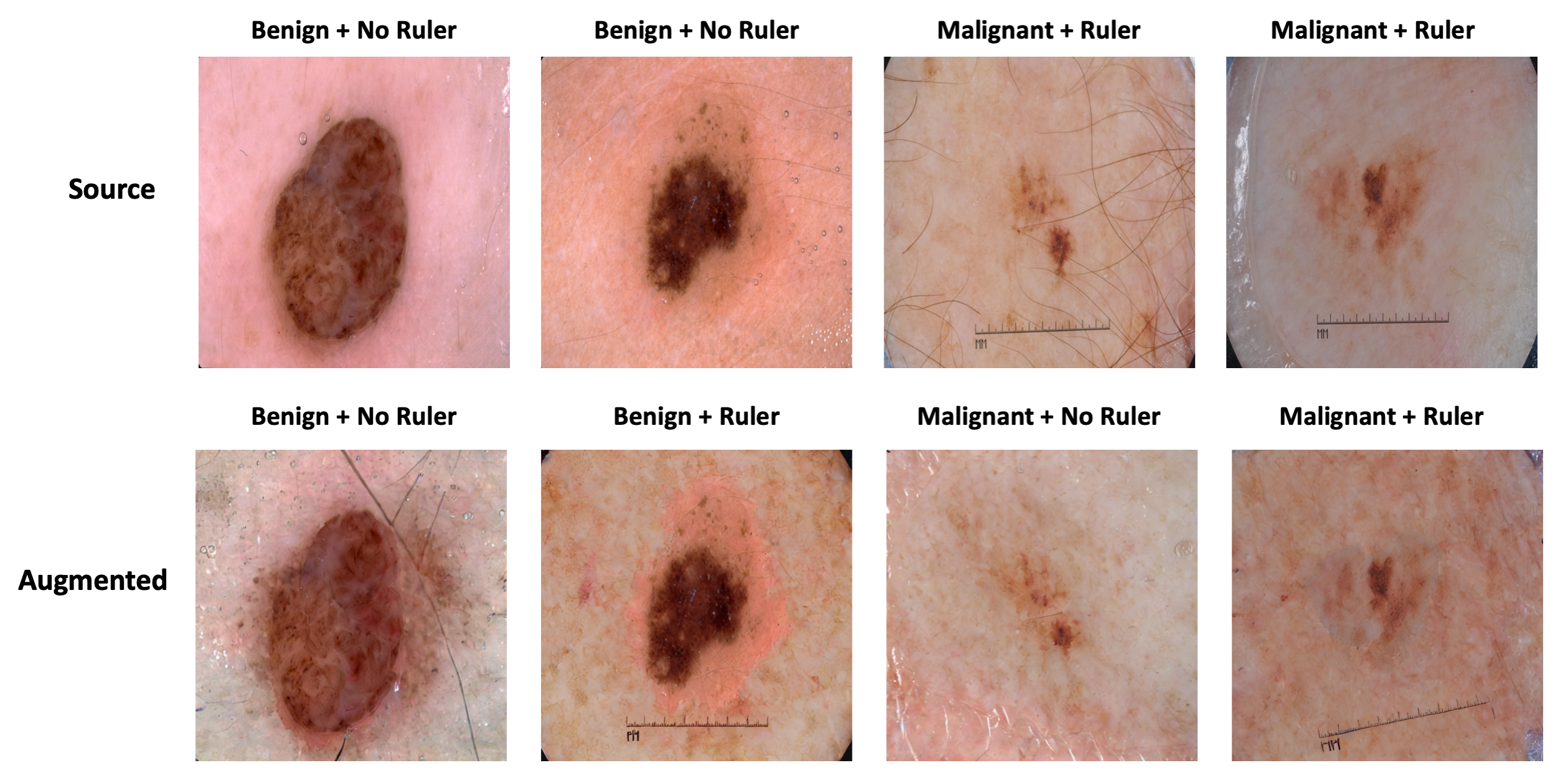}

    \caption{\textbf{MaskMedPaint Image Generation for ISIC 2018 Dermoscopic Shift.} All possible combinations of skin lesion conditions and rulers in the generated augmentations. 
    }
    \label{fig:isic_image_gen}
\end{figure}

In the ISIC 2018 dataset, the artifact of surgical ruler marking is spuriously correlated with the presence of melanoma. 
From Table \ref{table:medical_results}, Masked baseline achieves the best target domain performance, with overlapping CIs with the second-best MaskedMedPaint. The other baseline methods fail to mitigate the extreme bias, as exemplified by high source accuracies above 0.9 and low target accuracies below 0.2. To provide a reference, we trained a best-possible ``oracle'' classifier on an expanded ISIC dataset with artifact annotations (840 source images, 1031 target images). The oracle achieves a source accuracy of 0.628 and a target accuracy of 0.742. We see that both MaskMedPaint and LADS perform more similarly to the more ideal oracle classifier not dependent on rulers as a spurious feature. Lastly, we trained a ``ground-truth'' classifier to predict the presence of rulers (Accuracy 0.985) in the generated images. We find that MaskMedPaint adds the rulers for 39.3\% of the benign images, and removes the rulers for 46.9\% of the malignant images.

\subsection{Generalizing Dataset Shifts in Chest X-ray Images}
\begin{figure}[ht!]
    \centering
    \includegraphics[width=\linewidth, valign=m]{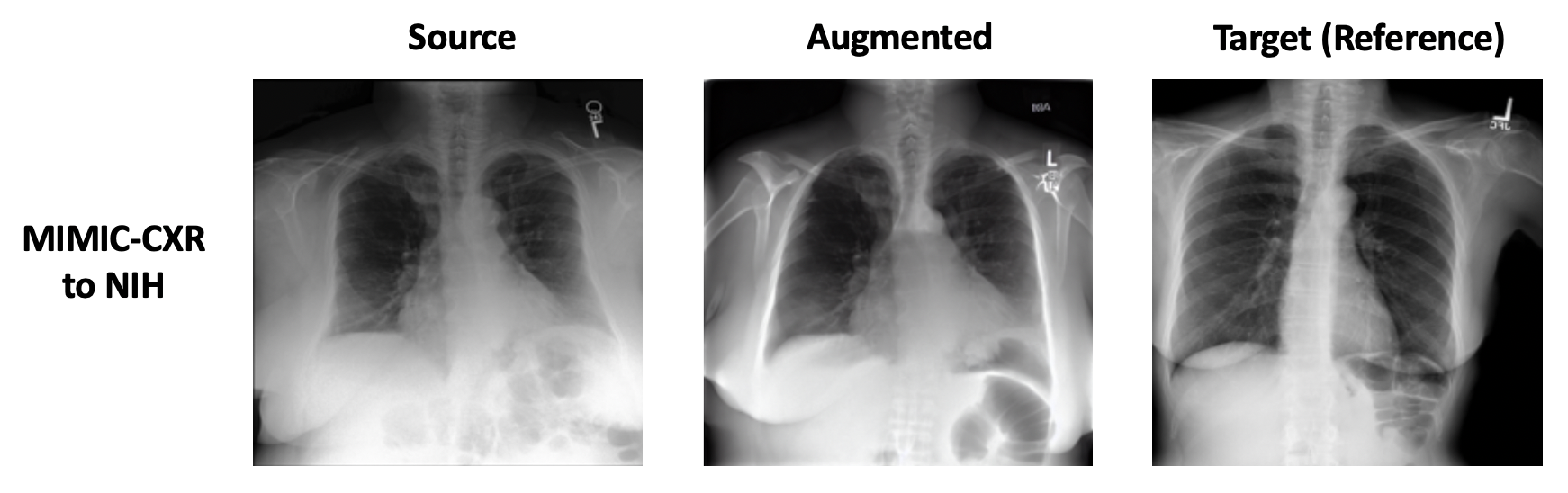}

    \caption{\textbf{MaskMedPaint Image Generation for CXR dataset shift.} Example of the source MIMIC-CXR image (left) augmented to NIH style with MaskMedPaint (middle). For reference, a CXR from NIH (right).}
    \label{fig:cxr_gen}
\end{figure}

We evaluate global shifts in a medical setting with a transfer between Chest X-ray datasets. With a transfer from MIMIC-CXR to NIH ChestXray14, we see from Table \ref{table:medical_results} that MaskedMedPaint has the best generalization to target domain (AUROC 0.546) with LADS being a close second (AUROC 0.521). All methods except Masked perform with higher than AUROC 0.7 in the source domain. Qualitatively, we see that artifacts such as radiology markers and general image contrast can differ in the generated augmentations (Figure \ref{fig:cxr_gen}). 

\section{Conclusion}
In this paper, we propose MaskMedPaint, a data augmentation method for mitigating spurious correlation through leveraging text-to-image diffusion models. 
Our method is promising for medical imaging applications, in which the spurious features are hard to describe in natural language, and data samples from the actual target domain are often limited in size or unlabeled. Our framework offers exciting possibilities for future research, such as spurious feature discovery from augmented images and a controlled generation of spurious features in particular spatial locations.

\newpage
\bibliographystyle{unsrtnat}
\bibliography{jmlr-sample}

\appendix

\section{Appendix}
\subsection{Dataset Splits and Preprocessing Details}
\label{appendix:dataset}

\begin{table}[H]
\begin{center}
 \begin{tabular}{lllll}
 \toprule
\multirow{2}{*}{} & \multicolumn{2}{l}{Landbird} & \multicolumn{2}{l}{Waterbird} \\
 & land & water & land & water \\
 \midrule
Train & 770 & - & - & 230 \\
Val & 443 & - & - & 131 \\
Extra & - & 50 & 50 & - \\
Test & 770 & 230 & 770 & 230 \\
\bottomrule
\end{tabular}   
\end{center}
\caption{Dataset splits for Waterbirds}
\label{table:waterbird_dataset_split}
\end{table}
\begin{table}[H]
\begin{center}
\begin{tabular}{llll}
 \toprule
Shift & Split & Color & Grayscale \\
\midrule
grayscale-to-color & Train & - & 921 \\
 & Val & - & 50 \\
 & Extra & 100 & - \\
 & Test & 100 & 100 \\
 \bottomrule
\end{tabular}
\end{center}
\caption{Dataset splits for iWildcam color shift.}
\label{table:waterbird_dataset_split_color}
\end{table}
\begin{table}[H]
\begin{center}
\begin{tabular}{llll}
\toprule
Shift & Split & Day & Night \\
\midrule
night-to-day & Train & - & 516 \\
 & Val & - & 50 \\
 & Extra & 100 & - \\
 & Test & 100 & 100 \\
 \bottomrule
\end{tabular}
\end{center}
\caption{Dataset splits for iWildcam time shift.}
\label{table:waterbird_dataset_split_time}
\end{table}
\begin{table}[H]
\begin{center}
\begin{tabular}{lllll}
\toprule
 & \multicolumn{2}{l}{Malignant} & \multicolumn{2}{l}{Benign} \\
 & ruler & w/o ruler & ruler & w/o ruler \\
 \midrule
Train & 228 & - & - & 811 \\
Val & 10 & - & - & 40 \\
Extra & - & 50 & 50 & - \\
Test & 70 & 70 & 70 & 70 \\
\bottomrule
\end{tabular}
\end{center}
\caption{Dataset splits for ISIC 2018 Dermoscopic Images.}
\label{table:isic_dataset_split}
\end{table}
\begin{table}[H]
\centering
\begin{tabular}{lll}
\toprule
Split & MIMIC-CXR & NIH \\
\midrule
Train & 1408 & - \\
Val & 219 & - \\
Extra & - & 100 \\
Test & 422 & 405 \\
\bottomrule
\end{tabular}
\caption{Dataset splits for CXR images}
\label{table:cxr_dataset_split}
\end{table}

\paragraph{Waterbirds}
Waterbirds \citep{sagawa2019distributionally} is a dataset for bird classification with spurious background bias. Similar to \citep{dunlap2024diversify}, we construct a version of Waterbirds such that the train and validation sets contain only the source groups, the extra dataset contains only target groups, and the test dataset is balanced. In our diffusion pipeline, we assume access to the class label (``landbird'' vs ``waterbird'') in the source images. 

\paragraph{iWildCam} 
The iWildCam 2020 dataset \citep{koh2021wilds} consists of heat or motion-activated camera-trap images of wildlife for the classification of animal species. Images across different camera-trap locations can vary greatly in background, vegetation, color, illumination, and camera angles. Since the frequency of animal species vary across locations, location-specific attributes can serve as spurious shortcuts in classifier learning. Similar to \citep{dunlap2024diversify}, we focus on the prediction of 6 animal classes (cattle, elephant, impala, zebra, giraffe, dik-dik). Unlike their general dataset split, however, we construct our dataset such that we can investigate the following shifts in more controlled settings: 1) \textbf{Color shift:} grayscale images to color images across different traps with similar backgrounds and 2) \textbf{Time shift:}  nighttime images to daytime images for grayscale traps.  

\paragraph{ISIC Dermoscopy}
We aim to construct an analogous dataset to Waterbirds in the medical setting. International Skin Imaging Collaboration (ISIC) 2018 \citep{codella2019skin} is a dataset of dermoscopic skin images that are labeled for skin conditions such as melanoma. Using lesion and artifact annotations provided by \citep{bissoto2020debiasing}, we construct a dataset such that the skin lesion images with melanoma are spuriously associated with a background spurious feature of ruler-like markings (see Figure \ref{fig:isic_image_gen}). We follow a similar dataset split to Waterbirds, with train and validation datasets being the source (e.g. melanoma with ruler, no melanoma without ruler), the extra dataset being target (e.g. melanoma without a ruler, no melanoma with a ruler), and the test dataset being balanced. We use the ruler annotations from \citep{bissoto19deconstructing}. We further preprocessed this split ($n=2594$) by removing images that had a label other than ``benign" or ``malignant", with 2587 images remaining. We filtered out an additional 187 images that contained patches, as this provided a separate strong spurious signal aside from the ruler markings. 

\paragraph{Chest X-rays}
MIMIC-CXR-JPG \citep{johnson2019mimic} is a collection of chest radiographs from Beth Israel Deaconess Medical Center between the years 2011 to 2016. NIH ChestXray14 \citep{wang2017hospital} is a collection of chest radiographs provided by the NIH Clinical Center. We standardize the labels across the two datasets for the conditions of Atelectasis, Cardiomegaly, Edema, Pneumothorax, and No Finding. We subset to radiographs of the posteroanterior (PA) view. All performance metrics are averaged over the 5 predictive classes. The shift that we evaluate is a general shift between the datasets in a transfer from MIMIC-CXR to NIH. 

\subsection{Additional Implementation Details}
\label{appendix:additional_implementation}
\paragraph{Step 1. Class-Conditional Fine-tuning on Source Images} To fine-tune Stable Diffusion on the source dataset, we use hyperparameters of learning rate=$1.0 \mathrm{e}-5$, constant lr scheduler, snr\_gamma=5.0, resolution=512, max\_train\_steps=2500, and checkpointing\_steps=500. We run the ``train\_text\_to\_image.py'' script from the diffusers library. 

\paragraph{Step 2. Unconditional Fine-tuning on Target Images} To fine-tune the Dreambooth model, we use hyperparameters of learning rate=$1.0 \mathrm{e}-6$, constant lr scheduler, snr\_gamma=5.0, resolution=512, max\_train\_steps=2500, and checkpointing\_steps=500. We run the ``train\_dreambooth.py'' script from the diffusers library.

\subsection{Mitigating Background Bias in Waterbirds}
\begin{figure}[ht!]
    \centering
    \includegraphics[width=\linewidth]{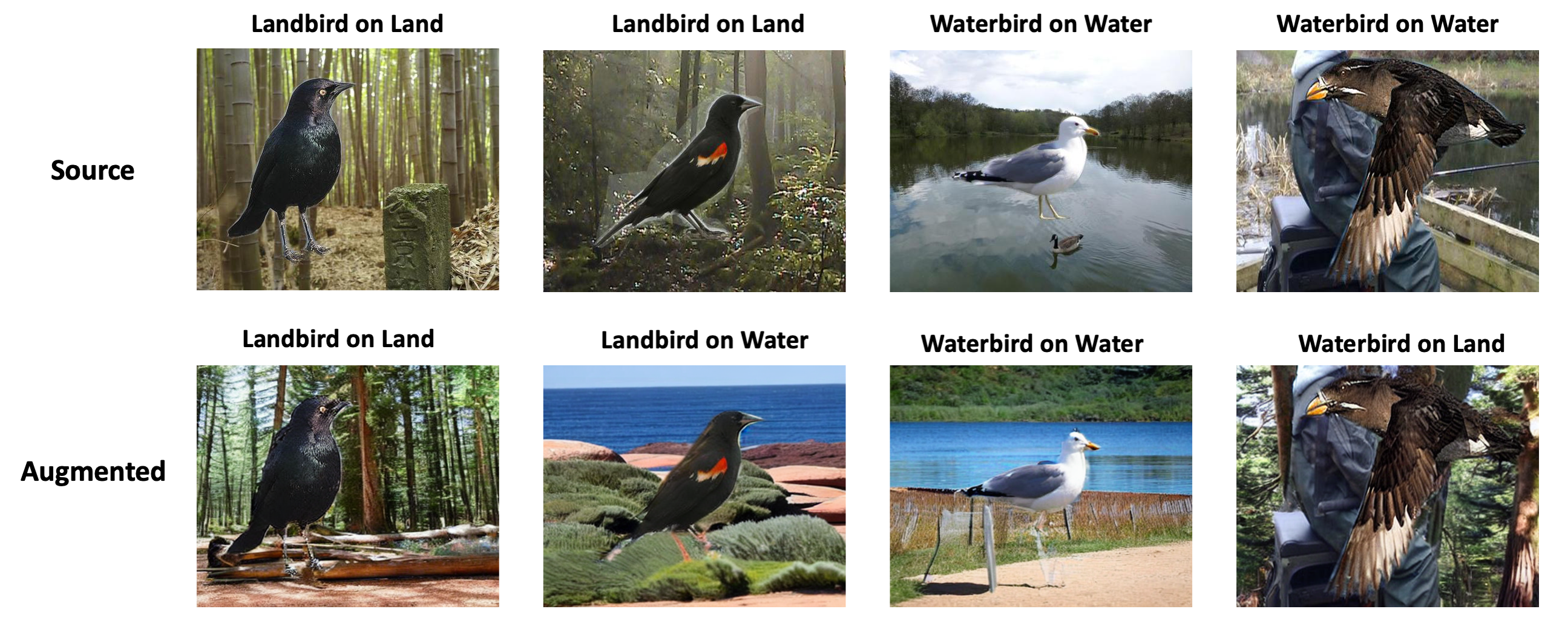}
    \caption{\textbf{MaskMedPaint Image Generation for Waterbirds Shift.} All possible combinations of bird species and backgrounds in the generated augmentations. Landbirds on water and waterbirds on land are categories not present in the original source dataset and are counterfactuals generated by MaskMedPaint. }
    \label{fig:waterbirds_image_gen}
\end{figure}

Initially, we evaluate the ability of MaskMedPaint in the Waterbirds dataset. The background is spuriously associated with the predicted bird species.
As shown in Table \ref{table:natural_results}, LADS performs the best for both source and target domains. If we do not assume access to a pretrained CLIP model, however, MaskMedPaint provides the best generation to the target domain with an accuracy of 0.571. Qualitatively, generated images are reasonable - as shown in Figure \ref{fig:waterbirds_image_gen}, the bird is both well-segmented and transferred into new land or water backgrounds that match the style of the unlabeled target images the diffusion model was finetuned on. With a ``ground-truth'' classifier trained to predict both the bird and background classes (Accuracy 0.953),  we find that 59.8\% of source waterbird images are augmented to land backgrounds and 37.2\% of landbird images are augmented to water backgrounds. Thus, although not every generated image is a counterfactual, this proportion is enough for the augmented classifier to learn to focus on the bird instead of the background.

\begin{table*}[t]
\centering
\begin{tabular}{ll|rrr|rrr}
\toprule
\textbf{Dataset} & \textbf{Method} & \multicolumn{3}{c}{\textbf{Source Test Accuracy}} & \multicolumn{3}{c}{\textbf{Target Test Accuracy}} \\
\midrule
\multirow{7}{*}{Waterbirds} & \multicolumn{1}{c}{\textbf{}} & \multicolumn{1}{c}{\textbf{Mean}} & \multicolumn{1}{c}{\textbf{\begin{tabular}[c]{@{}c@{}}Lower \\ 95 CI\end{tabular}}} & \multicolumn{1}{c}{\textbf{\begin{tabular}[c]{@{}c@{}}Upper \\ 95 CI\end{tabular}}} & \multicolumn{1}{c}{\textbf{Mean}} & \multicolumn{1}{c}{\textbf{\begin{tabular}[c]{@{}c@{}}Lower \\ 95 CI\end{tabular}}} & \multicolumn{1}{c}{\textbf{\begin{tabular}[c]{@{}c@{}}Upper \\ 95 CI\end{tabular}}} \\
\midrule
 & Base & 0.916 & 0.906 & 0.925 & 0.264 & 0.228 & 0.300 \\
 & CutMix & \multicolumn{1}{l}{0.928} & 0.922 & 0.934 & 0.306 & 0.248 & 0.363 \\
 & Mixup & \multicolumn{1}{l}{0.918} & 0.913 & 0.923 & 0.320 & 0.276 & 0.363 \\
 & Masked & 0.750 & 0.560 & 0.941 & 0.310 & 0.147 & 0.472 \\
 & ALIA* & 0.939 & 0.937 & 0.942 & 0.250 & 0.176 & 0.325 \\
 & LADS* & {\ul \textbf{0.967}} & {\ul \textbf{0.965}} & {\ul \textbf{0.970}} & {\ul \textbf{0.732}} & {\ul \textbf{0.725}} & {\ul \textbf{0.740}} \\
 & MaskMedPaint & 0.919 & 0.898 & 0.941 & 0.571 & 0.456 & 0.686 \\
 \midrule
\multirow{7}{*}{\begin{tabular}[c]{@{}l@{}}iWildCam\\ (Grayscale to Color)\end{tabular}} & Base & \textbf{0.980} & \textbf{0.973} & \textbf{0.988} & 0.286 & 0.228 & 0.344 \\
 & CutMix & {\ul \textbf{0.985}} & {\ul \textbf{0.969}} & {\ul \textbf{1.000}} & 0.210 & 0.162 & 0.257 \\
 & Mixup & 0.959 & 0.921 & 0.998 & 0.222 & 0.153 & 0.291 \\
 & Masked & 0.280 & 0.239 & 0.321 & 0.202 & 0.175 & 0.229 \\
 & ALIA* & 0.979 & 0.965 & 0.992 & 0.254 & 0.201 & 0.308 \\
 & LADS* & 0.952 & 0.936 & 0.968 & {\ul \textbf{0.508}} & {\ul \textbf{0.475}} & {\ul \textbf{0.541}} \\
 & MaskMedPaint & 0.969 & 0.942 & 0.997 & \textbf{0.399} & \textbf{0.248} & \textbf{0.550} \\
 \midrule
\multirow{7}{*}{\begin{tabular}[c]{@{}l@{}}iWildCam\\ (Night to Day)\end{tabular}} & Base & {\ul \textbf{0.971}} & {\ul \textbf{0.956}} & {\ul \textbf{0.986}} & 0.170 & 0.131 & 0.209 \\
 & CutMix & 0.695 & 0.349 & 1.041 & 0.199 & 0.122 & 0.276 \\
 & Mixup & 0.643 & 0.261 & 1.025 & 0.179 & 0.096 & 0.262 \\
 & Masked & 0.324 & 0.283 & 0.365 & 0.174 & 0.081 & 0.268 \\
 & ALIA* & \textbf{0.929} & \textbf{0.876} & \textbf{0.982} & 0.155 & 0.130 & 0.179 \\
 & LADS* & 0.858 & 0.825 & 0.89 & 0.204 & 0.164 & 0.244 \\
 & MaskMedPaint & 0.928 & 0.905 & 0.952 & {\ul \textbf{0.333}} & {\ul \textbf{0.306}} & {\ul  \textbf{0.361}}\\
\bottomrule
\end{tabular}
\caption{Performance metrics evaluated across the source and target domains for the natural image datasets. Best-performing method is bolded and underlined, and the second-best is bolded if its CI overlaps with the CI of the best-performing. Methods requiring additional pretrained models are highlighted with *.}
\label{table:natural_results}
\end{table*}
\subsection{Generalizing Imaging Shifts in iWildCam} 
\begin{figure}[ht!]
    \centering
    \includegraphics[width=\linewidth, valign=m]{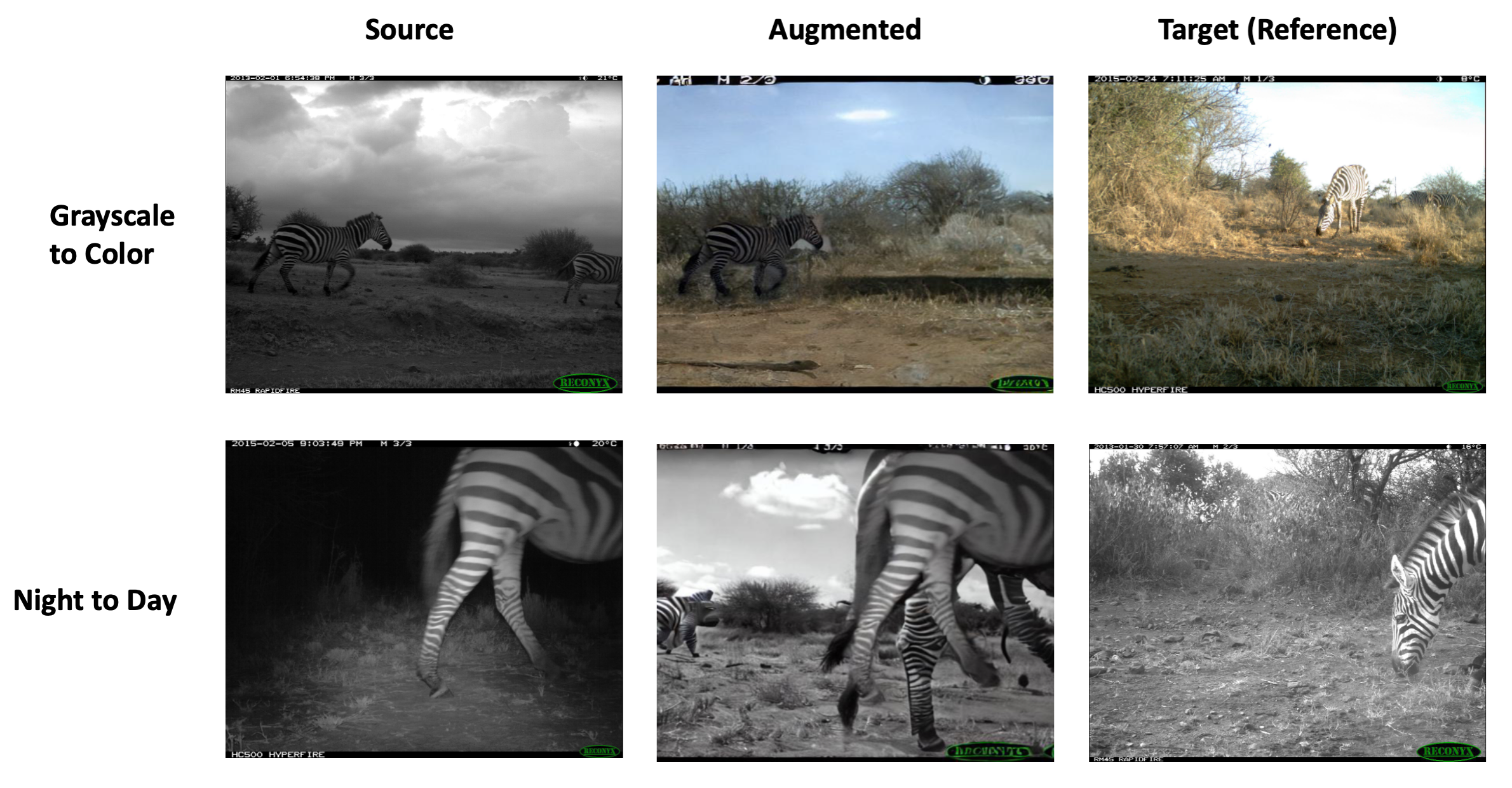}

    \caption{\textbf{MaskMedPaint Image Generation for iWildcam Shifts.} Example of the original source image (left) adapted to the style of the target domain through MaskMedPaint augmentation (middle). For reference, we provide the real target domain images from the same class (right).}
    \label{fig:iwildcam_image_gen}
\end{figure}

We also identify the extent to which our method can mitigate bias for global shifts in natural images. 
In the iWildCam dataset, we evaluate two common shifts: Grayscale to Color Shift (camera trap variation) and Night to Day Shift (time of day variation).

For Grayscale to Color shift, LADS performs the best in the target domain, with overlapping CIs with the second-best MaskMedPaint. For Night to Day Shift, MaskMedPaint performs significantly better than the other baselines in the target domain.  
In the source domain, all methods except for Masked perform comparably. We hypothesize that since animals are often occluded or camouflaged with the environment, the segmentation model may struggle to identify the animal, leading to issues with the Masked baseline as all regions outside the segmented animal are masked during training. This issue in not present in MaskMedPaint since the entire image is still given to the diffusion model during the inpainting stage.

Qualitatively, MaskedMedPaint consistently converts grayscale images to colored versions and converts nighttime images to daytime versions. We do observe slight generalization shifts, however, as the augmented colored images may appear less saturated than natural colored images, and the augmented daytime images might appear darker or have dark borders (Figure \ref{fig:iwildcam_image_gen}). We believe further investigation into the diffusion process can potentially reduce this gap. 

\subsection{Ablations}
\textbf{Number of Real vs. Generated Images Added}
\begin{figure}[ht!]
    \centering
    \begin{tabular}{cc}
    \includegraphics[width=.45\linewidth,valign=m]{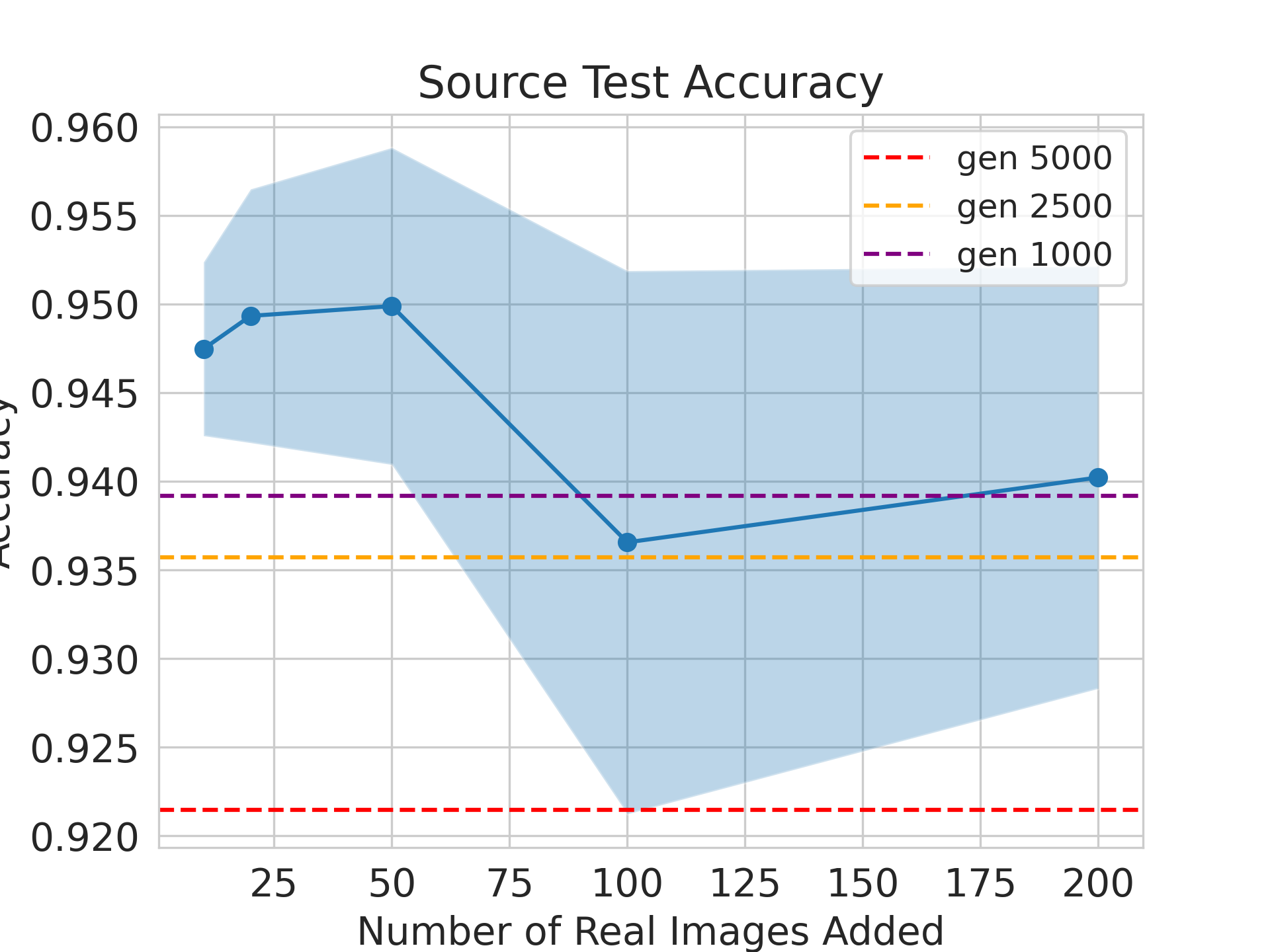} & \includegraphics[width=.45\linewidth,valign=m]{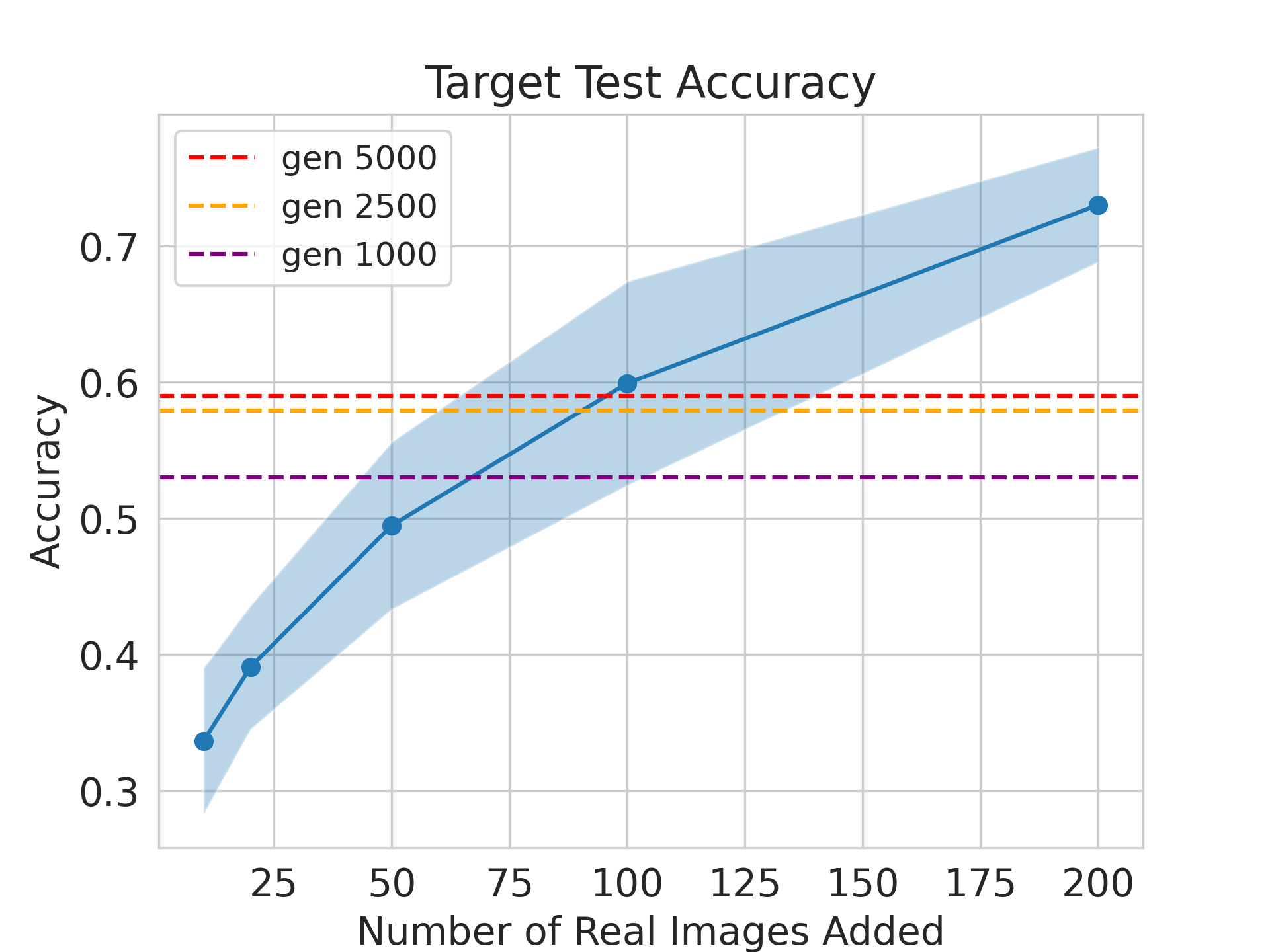} \\
\end{tabular}

    \caption{\textbf{Number of Real versus Generated Images Added (Waterbirds).} The overall test accuracy (left), source accuracy (middle), and target accuracy (right) of adding 10, 20, 50, 100, and 200 real images from the target distribution. As dashed lines, we have the mean accuracy of adding 1000 (purple), 2500 (yellow), and 5000 (red) MaskMedPaint generated images. CIs over 5 seeds.}
    \label{fig:waterbirds_ablation}
\end{figure}

To understand the number of generated images needed to approximate the performance improvement of adding real target images to the training data, we evaluated the Waterbirds classifier over an ablation of adding 10, 20, 50, 100, and 200 real images, in comparison with adding 1000, 2500, and 5000 generated images from MaskMedPaint. We see from Figure \ref{fig:waterbirds_ablation} that in terms of overall and target accuracy, adding 2500 generated images results in a similar performance to adding 100 real images. Doubling the amount of generated images to 5000 offers marginal improvement in target accuracy, possibly indicating a plateau of performance gain possible through adding strictly generated images from a single diffusion model. 

\textbf{Comparison of Diffusion-based Image Generation}

\begin{figure}[ht!]
    \centering
    \begin{tabular}{cc} \includegraphics[width=.45\linewidth,valign=m]{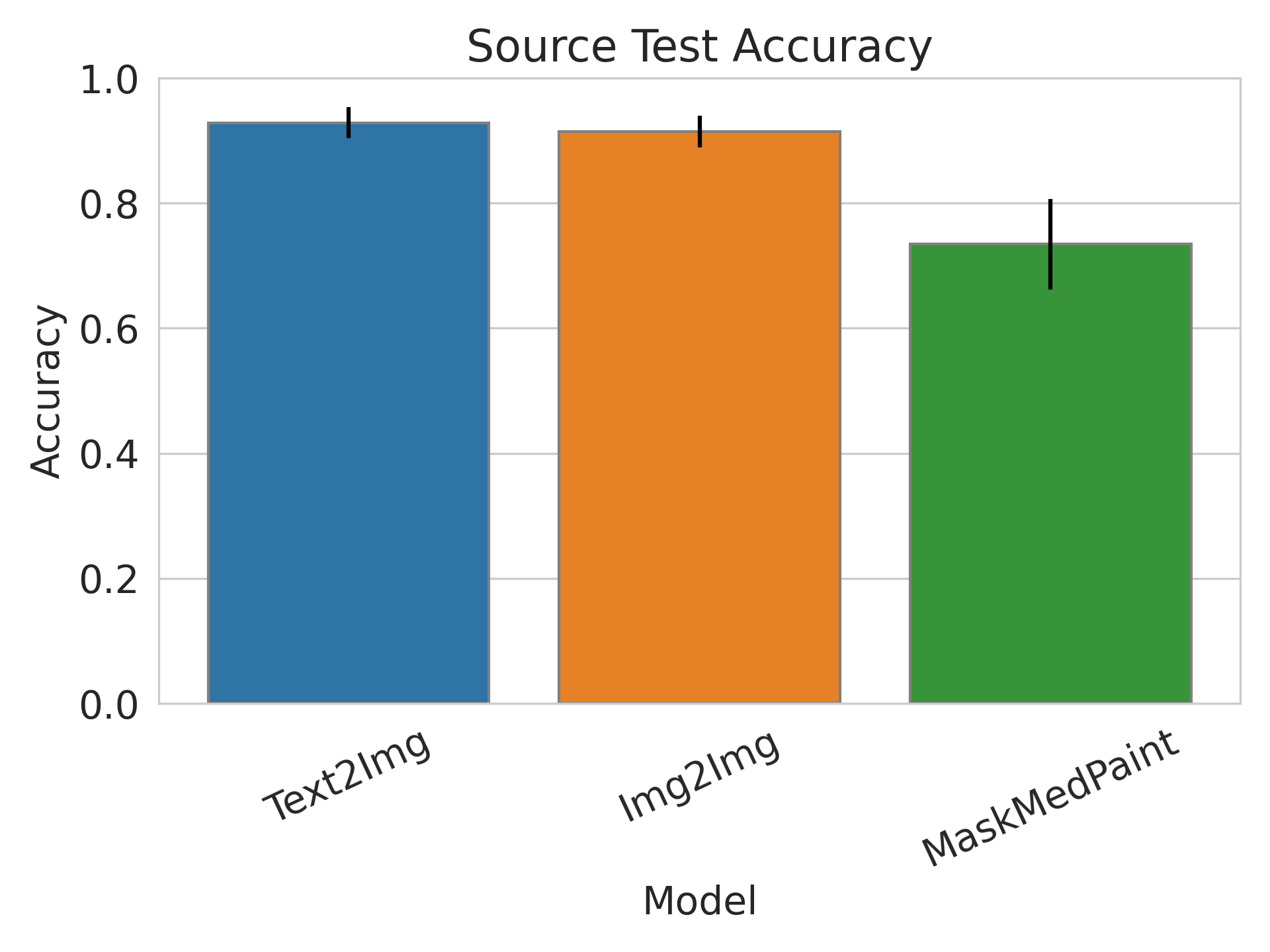} & \includegraphics[width=.45\linewidth,valign=m]{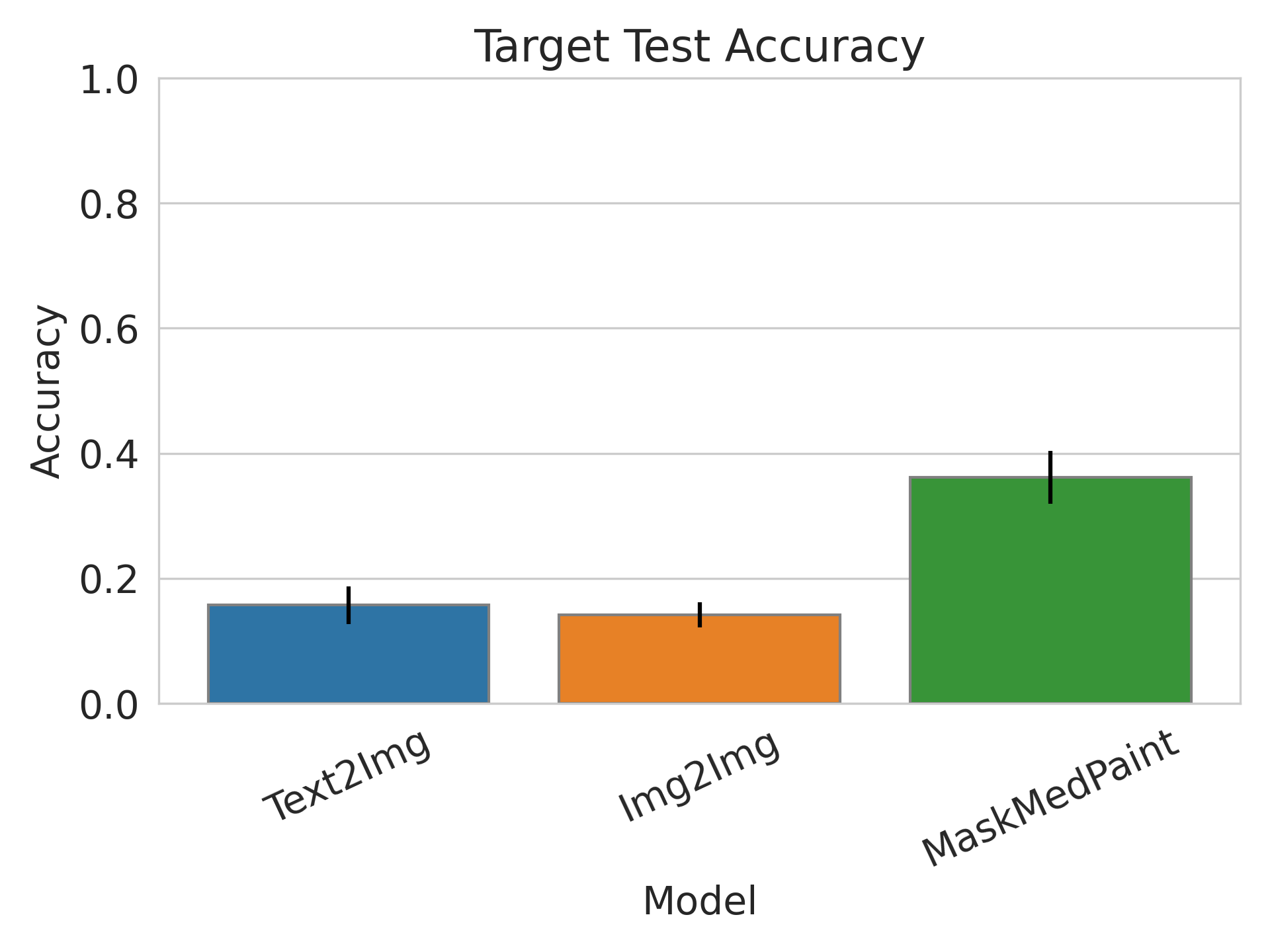} \\
\end{tabular}

    \caption{\textbf{Different Image Generation Methods (ISIC).} The overall test accuracy (left), source accuracy (middle), and target accuracy (right) of our MaskMedPaint method and the baselines of Text2Img and Img2Img generation. CIs over 5 seeds.}
    \label{fig:isic_ablation}
\end{figure}
In this section, we compare Inpainting with other diffusion-based image editing methods. Text2Img is the standard pipeline in which only the text prompt is used to generate images. Img2Img is a modified pipeline in which an initial image is used in addition to the text prompt. In the ISIC dataset, Text2Img and Img2Img both fail to correct for the spurious dependency (Fig. \ref{fig:isic_ablation}). Source test accuracies are falsely high, while target accuracies are low. Thus, masking the region of interest is essential for preserving features crucial for class predictions, especially in medical settings. 

\subsection{Limitations} 
While MaskMedPaint works well for our defined settings, it has certain limitations. First, we assume that the spurious feature can be separated from the region of interest identified by the segmentation model. If the main spurious features overlap with the non-inpainted region, the diffusion model may not generate effective counterfactuals that break the spurious correlation. 
Second, our method assumes access to approximately 100 unlabeled target domain images. In our preliminary experiments, we find that using only 10 to 20 target images leads to suboptimal results. The model tends to memorize the target images during the Dreambooth-based fine-tuning, reducing generation diversity. Lastly, we emphasize that real images are preferable to generated ones whenever possible. In the medical context, prioritizing the collection of more diverse data is crucial to effectively address bias rather than relying solely on data augmentation techniques. We do not advocate for the out-of-the-box use of our model to rectify gender or racial biases in image data, as the relationship between sensitive attributes and disease presentation is complex and requires careful, manual handling. 
\subsection{Text Prompts used for Stable Diffusion Fine-tuning and Inference}
\label{appendix:prompt}
\begin{table*}[htbp]
\begin{tabular}{llll}
\toprule
\textbf{Dataset} & \textbf{Pipeline Step} & \textbf{Prompt} & \textbf{Possible Tokens} \\
\midrule
\multirow{3}{*}{Waterbirds} & \begin{tabular}[c]{@{}l@{}}Source \\ Finetuning\end{tabular} & ``a photo of {[}CLASS{]}`` & ``landbird``, ``waterbird`` \\
 & \begin{tabular}[c]{@{}l@{}}Target \\ Finetuning\end{tabular} & ``a photo of {[}DUMMY{]}`` & ``target background`` \\
 & Inference & \begin{tabular}[c]{@{}l@{}}``a photo of {[}CLASS{]} \\ with {[}DUMMY{]}``\end{tabular} & Each from above \\
\multirow{3}{*}{iWildCam} & \begin{tabular}[c]{@{}l@{}}Source \\ Finetuning\end{tabular} & \begin{tabular}[c]{@{}l@{}}``a camera trap photo\\  of {[}CLASS{]}``\end{tabular} & \begin{tabular}[c]{@{}l@{}}``cattle``, ``elephants``, \\ ``impalas``, ``zebras``, \\ ``giraffes``, ``dik-diks``\end{tabular} \\
 & \begin{tabular}[c]{@{}l@{}}Target \\ Finetuning\end{tabular} & \begin{tabular}[c]{@{}l@{}}``a camera trap photo \\ with {[}DUMMY{]}``\end{tabular} & ``target-domain`` \\
 & Inference & \begin{tabular}[c]{@{}l@{}}``a camera trap photo of\\  {[}CLASS{]} with {[}DUMMY{]}``\end{tabular} & Each from above \\
\multirow{3}{*}{ISIC} & \begin{tabular}[c]{@{}l@{}}Source \\ Finetuning\end{tabular} & \begin{tabular}[c]{@{}l@{}}``a dermoscopic image of\\  {[}CLASS{]} skin lesion``\end{tabular} & ``benign``, ``malignant`` \\
 & \begin{tabular}[c]{@{}l@{}}Target \\ Finetuning\end{tabular} & \begin{tabular}[c]{@{}l@{}}``a dermoscopic image of\\  {[}DUMMY{]} skin lesion``\end{tabular} & ``target`` \\
 & Inference & \begin{tabular}[c]{@{}l@{}}``a dermoscopic image of\\  {[}CLASS{]}-{[}DUMMY{]} skin lesion``\end{tabular} & Each from above \\
 \multirow{3}{*}{CXR} & \begin{tabular}[c]{@{}l@{}}Source \\ Finetuning\end{tabular} & \begin{tabular}[c]{@{}l@{}}``a radiograph from dataset \\ {[}SOURCE{]} with conditions {[}CLASS{]} \end{tabular} &  
 \begin{tabular}[c]{@{}l@{}}``Atelectasis'', ``Cardiomegaly'', \\ ``Edema'', ``Pneumothorax'', \\ ``No Finding''\end{tabular} \\
 & \begin{tabular}[c]{@{}l@{}}Target \\ Finetuning\end{tabular} & \begin{tabular}[c]{@{}l@{}}``a radiograph from dataset \\  {[}DUMMY{]}``\end{tabular} & ``target`` \\
 & Inference & \begin{tabular}[c]{@{}l@{}}``a radiograph from dataset\\  {[}DUMMY{]} with conditions {[}CLASS{]} ``\end{tabular} & Each from above \\
 \bottomrule
\end{tabular}
\caption{The specific text prompt templates used in the MaskMedPaint pipeline.}
\label{table:prompt}
\end{table*}

\end{document}